\pdfoutput=1
\pdfmapfile{+cm.map}

\documentclass[11pt, a4paper]{custom_class}

\usepackage{custom_style}
\usepackage[utf8]{inputenc} 
\usepackage[T1]{fontenc}    
\usepackage{hyperref}       
\usepackage{url}            
\usepackage{booktabs}       
\usepackage{amsfonts}       
\usepackage{nicefrac}       
\usepackage{microtype}      
\usepackage{cleveref}       
\usepackage{lipsum}         
\usepackage{graphicx}
\usepackage{doi}
\usepackage{glossaries}
\usepackage{natbib}
\title{Philosophy of Cognitive Science\\In the Age of Deep Learning}
\author{
Raphaël Millière \\
Department of Philosophy \\
Macquarie University \\
\texttt{raphael.milliere@mq.edu.eu} \\
}

\renewcommand{\headeright}{}
\renewcommand{\undertitle}{\textcolor{red}{Forthcoming in \emph{WIREs Cognitive Science}}}
\renewcommand{\shorttitle}{Philosophy of Cognitive Science in the Age of Deep Learning}

\date{}

\begin{document}
\maketitle
\begin{abstract}
Deep learning has enabled major advances across most areas of artificial
intelligence research. This remarkable progress extends beyond mere
engineering achievements and holds significant relevance for the
philosophy of cognitive science. Deep neural networks have made
significant strides in overcoming the limitations of older connectionist
models that once occupied the centre stage of philosophical debates
about cognition. This development is directly relevant to long-standing
theoretical debates in the philosophy of cognitive science. Furthermore,
ongoing methodological challenges related to the comparative evaluation
of deep neural networks stand to benefit greatly from interdisciplinary
collaboration with philosophy and cognitive science. The time is ripe
for philosophers to explore foundational issues related to deep learning
and cognition; this perspective paper surveys key areas where their
contributions can be especially fruitful.
\end{abstract}

\hypertarget{sec-intro}{%
\section{Introduction}\label{sec-intro}}

Deep learning has enabled major breakthroughs in virtually every area of
artificial intelligence over the past decade -- including computer
vision, game playing, robotics, speech recognition, and natural language
processing. In most of these domains, deep neural networks (DNNs) have
matched or exceeded human performance on long-standing challenges. For
example, DNNs surpass humans on standard image classification benchmarks
\citep{heDeepResidualLearning2016}, beat world champions at chess and Go
\citep{silverMasteringGameGo2016, silverMasteringChessShogi2017},
achieve top scores on many tests including medical and law exams
\citep{openaiGPT4TechnicalReport2023}, and generate text often
indistinguishable from human writing
\citep{jonesDoesGPT4Pass2023, schwitzgebelCreatingLargeLanguage2024}.

The history of artificial neural networks is deeply interwined with
theoretical and empirical research exploring their adequacy as
computational models of human cognition. For much of its history, this
research programme yielded only modest empirical results, with neural
networks often comparing disfavorably to concurrent symbolic approaches.
The recent achievements of DNNs on real-world challenges stand in stark
contrast to the limited success of older neural network models. Yet they
are often perceived as mere engineering feats, increasingly enabled by
product-oriented research from technology companies rather than
academia.

While the main focus of deep learning research may not be on
understanding human cognition, it would be naive to discount the
potential contribution of engineering advances to scientific research.
Technical breakthroughs can open up new research directions that
catalyze scientific progress, sometimes in unexpected ways. Academic
researchers have been quick to leverage advances in deep learning to
build better models in computational linguistics, psychology, and
neuroscience. In return, deep learning researchers are increasingly
borrowing from the interdisciplinary toolkit of cognitive science -- if
not to build models, at least to evaluate them.

This calls for a reappraisal of the place of modern artificial neural
networks within the project of cognitive science. What is the relevance
of the progress of deep learning for cognitive science? Conversely, what
is the relevance of cognitive science to deep learning research? This
paper will provide an opinionated perspective on these questions through
the lens of the philosophy of cognitive science.\footnote{While I will
  consider the progress of deep learning as a whole in relation to
  cognitive science, I will pay particular attention to large language
  models (LLMs), whose capabilities have invited more specific
  comparisons with human cognition (see
  \citealt{millierePhilosophicalIntroductionLanguage2024} for discussion).}
Section~\ref{sec-coming-age-connectionism} provides a very brief outline
of the history of neural network research leading to deep learning
Section~\ref{sec-deep-learning-models-cognitive-models} examines whether
and how recent developments in deep learning can inform research on
human cognition. Finally, Section~\ref{sec-methodological-issues}
discusses how insights from cognitive science and philosophy can help
adress ongoing methodological issues with the evaluation of DNNs and
human-machine comparisons.

\hypertarget{sec-coming-age-connectionism}{%
\section{The coming of age of
connectionism}\label{sec-coming-age-connectionism}}

Cognitive science has always been centrally informed by theoretical
concepts from computer science
\citep{bodenMindMachineHistory2008, millerCognitiveRevolutionHistorical2003}.
In its early days, the dominant research programme sought to explain
human cognition through computations over structured symbolic
representations, analogous to rule-based programs executed by digital
computers \citep{newellComputerScienceEmpirical1976}. Artificial neural
networks (ANNs), also known as connectionist models, have come to play
an increasingly important role in challenging and transforming this
classical research programme, with a few significant milestones leading
to modern DNNs.

ANNs consist of simple neuron-like units connected into networks via
weighted connections. The units are segregated into an input layer that
receives data to be processed, an output layer to produce adequate
responses, and one or more hidden layers in between that learn to
represent features and perform computations that map inputs to outputs.
McCulloch and Pitts laid the groundwork for connectionism by introducing
a simplified mathematical model of neuron functioning, and proving that
networks of such neuron-like units can in principle compute any logical
function \citep{mccullochLogicalCalculusIdeas1943}. Building on this
pioneering work, Rosenblatt's perceptron later demonstrated that an ANN
with trainable connection weights could learn to classify patterns,
lending key support to connectionist explanations of learning
\citep{rosenblattPerceptronProbabilisticModel1958}.

Despite the theoretical dominance of the symbolic approach,
connectionism was revived in the 1980s with innovations such as
distributed representations, multilayer neural networks with hidden
layers allowing nonlinear decision boundaries, and the backpropagation
algorithm for training weights across multiple layers
\citep{rumelhartLearningRepresentationsBackpropagating1986a, rumelhartParallelDistributedProcessing1987}.
Connectionist models showed promise in explaining psychological
phenomena that had eluded traditional symbolic approaches, including
aspects of learning, memory, categorization, language, reasoning, and
vision
\citep{elmanFindingStructureTime1990, fukushimaNeocognitronSelforganizingNeural1980, kruschkeALCOVEExemplarbasedConnectionist1992, cohenControlAutomaticProcesses1990}.
These initial results challenged the assumption that the mind performs
serial computations over discrete symbolic representations.

The advent of deep learning from the mid-2000s onward enabled much
larger and more complex ANN architectures to be trained effectively
\citep{lecunDeepLearning2015}. This is due to a combination of factors
-- including better training techniques, the availability of much larger
datasets assembled from internet data, and major increases in
computational power. DNNs differ from older connectionist models in
several significant ways that account for their superior performance on
a broad range of challenging tasks
\citep{bucknerDeepLearningPhilosophical2019}. Their depth allows them to
hierarchically compose complex concepts from simpler features across
many hidden layers with increasing levels of abstraction. They also make
use of sophisticated architectures with heterogenous components to
promote specific inductive biases. For example, convolutional neural
networks used in computer vision have convolutional filters tuned to
detect specific features in images regardless of their position, while
Transformers used in LLMs have self-attention layers that track specific
dependencies between lexical items. Other tricks such as sparse
activations and regularization techniques prevent overfitting despite
the incredibly large number of trainable parameters in these networks.

Thanks to these innovations, DNNs are much better than previous neural
networks architectures at efficiently learning useful abstract
representations and inducing computations that can generalize beyond the
specific distribution of their training data. In this respect, they can
be seen as bringing the connectionist programme to fruition, by
demonstrating that neural networks that are no longer bounded by scarse
computing resources can overcome, in practice, many of the putative
theoretical limitations levelled against connectionism in the previous
decades.

\hypertarget{sec-deep-learning-models-cognitive-models}{%
\section{Deep learning models as cognitive
models}\label{sec-deep-learning-models-cognitive-models}}

The maturation of the connectionist programme in the guise of modern
DNNs is eminently relevant to several long-standing debates in (the
philosophy of) cognitive science.\footnote{While I will mainly focus on
  psychology here, DNNs also occupy a increasingly important place in
  cognitive neuroscience
  \citep[e.g.][]{richardsDeepLearningFramework2019, doerigNeuroconnectionistResearchProgramme2023, lindsayGroundingNeuroscienceBehavioral2024}.}
The most influential criticism of connectionism as a theory of cognition
came from proponents of the language of thought hypothesis, in the form
of a dilemma: either connectionist models are fundamentally inadequate
accounts of cognition, or they merely implement classical symbol
manipulation
\citep{fodorConnectionismCognitiveArchitecture1988a, fodorConnectionismProblemSystematicity1990, pinkerLanguageConnectionismAnalysis1988}.
If connectionist models are viewed as genuine alternatives to classical
architectures, classicists argued they fail to capture core
structure-sensitive properties of cognition, like productivity and
systematicity, because they lack genuinely compositional
representations. But if they can be viewed as implementations of
classical systems, then connectionists models are just showing how
symbols and rules could be realized in neural hardware. Either way,
classicists concluded that connectionist models come up short compared
to symbolic architectures in explaining core cognitive capacities like
language and reasoning. This dilemma has been revived in the age of deep
learning: if DNNs can match human performance on a broad spectrum of
perceptual and cognitive tasks, they must do so by implementing core
features of language of thought architectures
\citep{quilty-dunnBestGameTown2022, mandelbaumProblemsMysteriesMany2022, marcusDeepLearningCritical2018}.

Connectionists typically resist this dilemma in two ways. The first is
to suggest, on the basis of experimental results, that cognition is not
as regimented and systematic as classicists take it to be, and that
connectionist models should not be held to a higher standard than humans
themselves \citep{johnsonSystematicityLanguageThought2004}. The second
is to argue that connectionist models can in fact account for the
structure-sensitive properties of cognition and the constituent
structure of mental representations \emph{without} merely implementing a
classical architecture
\citep{smolenskyProperTreatmentConnectionism1988}. These two strategies
are not exclusive; together, they suggest that connectionist models and
human cognition can meet halfway between unstructured input-output
mapping and idealized systematicity.

In recent years, DNNs have moved much closer to bridging the gap with
human performance on structure-sensitive tasks. In particular, an
extensive body of work building on insights from cognitive science has
probed their capacity for systematic compositional generalization
\citep{donatelliCompositionalityComputationalLinguistics2023}. To rule
out confounds such as memorization of common compositional structures,
DNNs can be trained from scratch on synthetic datasets and evaluated on
held-out test samples
\citep[e.g.,][]{lakeGeneralizationSystematicityCompositional2018, kimCOGSCompositionalGeneralization2020}.
In these datasets, the train-test split is carefully designed such that
high accuracy on test samples requires systematically recombining
previously learned elements to map new inputs made up from these
elements to their correct output. This line of research has shown that
various parameters can have a significant impact on compositional
generalization in DNNs, including their architectural features and
training regime
\citep{csordasDevilDetailSimple2021, ontanonMakingTransformersSolve2022, qiuImprovingCompositionalGeneralization2022a, kazemnejadImpactPositionalEncoding2023}.
When these parameters are selected appropriately, DNNs can achieve good
performance on compositional generalization datasets without built-in
compositional rules.

For example, \citet{lakeHumanlikeSystematicGeneralization2023} show that
a standard Transformer-based neural network trained with meta-learning
can achieve human-like systematic generalization in a controlled
few-shot learning experiment, as well as state-of-the-art performance on
systematic generalization benchmarks. Their meta-learning approach
consists in training the network on a stream of artificial tasks, each
based on an underlying ``interpretation grammar'' that specifies
compositional mappings from instructions to output sequences. At test
time, the model achieves human-like accuracy and error patterns, without
the need for explicit compositional rules. While meta-learning from
different tasks helps promote compositional generalization, recent
research using a standard learning regime has also shown that simply
training a network past the point where it achieves excellent accuracy
on the training data can lead it to acquire more tree-structured
computations, and generalize significantly better to held-out test data
that require learning hierarchical rules
\citep{murtyGrokkingHierarchicalStructure2023}.

In line with the second horn of the classicist dilemma, one might
interpret these results as providing evidence that, given the right
architecture and training regime, modern DNNs can account for the
structure-sensitive properties of cognition by implementing a language
of thought \citep{quilty-dunnBestGameTown2022}. However, this conclusion
hinges on controversial assumptions about what the relevant notion of
implementation ought to be, and what kinds of properties should be taken
as specific evidence of implementing a language of thought
\citep{smolenskyConnectionismConstituentStructure1989, mcgrathPropertiesLoTsFootprints2023, pavlickSymbolsGroundingLarge2023}.
For example, one core feature of language of thought architectures is
the ability to perform variable binding over discrete symbolic
representations, where ``roles'' (variables) and ``fillers'' (values)
are represented independently. Mechanistic interpretability research
that seeks to reverse-engineer computations in trained DNNs does suggest
they can acquire a mechanism for variable binding
\citep{elhage2021mathematical, daviesDiscoveringVariableBinding2023, baroniProperRoleLinguistically2022, millierePhilosophicalIntroductionLanguage2024a}.
However, this mechanism implements a ``fuzzy'' form of variable binding
making use of vector subspaces that are not always functionally
equivalent to discrete memory slots \citep{olsson2022context};
accordingly, role-filler independence in these networks is not absolute,
but comes in degrees. This suggests that while modern DNNs can compute
over compositional representations with real constituent structure, this
structure is \emph{non-classical} and should not be taken to reflect the
core properties of a language of thought on pain of trivializing them.
For the language of thought hypothesis to remain substantive, it must
commit to specific claims about how representations are composed beyond
pointing to their constituent structure. However, committing to such
claims in light of the available evidence about modern DNNs may
undermine the second horn of the dilemma -- the view that their
successful behavior is best explained by positing that they merely
implement a language of thought architecture.

A related issue concerns the content-specificity of computations
performed by DNNs. Neural networks are traditionally assumed to learn
many specific input-output mappings. On this view, each layer-to-layer
transformation deals with a given input in a way that depends on the
particular content of that input, rather than general computational
principles applied across inputs. In other words, ANNs are generally
taken to perform only \emph{content-specific computations}, by contrast
with classical architectures. However, there is compelling evidence that
modern DNNs can also perform \emph{non-content-specific computations},
as argued by \citet{sheaMovingContentspecificComputation2023}. For
example, episodic deep reinforcement learning models apply
non-content-specific similarity computations to stored memory
representations. When a new state is encountered, the system retrieves
all previously stored memories and calculates their similarity to the
current state using the same similarity algorithm, regardless of what
the specific contents of those memories are. Transformer-based LLMs also
induce a broad repertoire of non-content-specific computations,
including domain-general ``induction head'' mechanisms that implement
the aforementioned capacity for variable binding
\citep{olsson2022context}.

One of the notable feature of DNNs, in contrast with classical systems,
is that their architecture does not delineate a strict distinction
between \emph{content-specific} and \emph{non-content-specific}
computations. Rather, as I previously alluded to when discussing
role-filler independence, the content-specificity of DNN computations is
a matter of degrees. This is significant if we take DNNs seriously as
cognitive models that provide genuine alternatives to classical
architectures, rather than mere implementations. For example, it
dovetails with recent findings about the behavioural convergence between
DNNs and humans on various classical reasoning tasks. Indeed, LLMs show
similar accuracy overall to humans on various reasoning problems,
including natural language inference, syllogism validity, or the Wason
selection task. Moreover, both humans and LLMs exhibit ``content
effects'' on reasoning tasks; that is, they tend to perform more
accurately when the content of a reasoning problem is familiar and
plausible \citep{dasguptaLanguageModelsShow2023}. The best LLMs also
match human performance across a range of verbal and non-verbal analogy
tasks requiring inductive reasoning about abstract relations, such as
Raven's progressive matrices or letter string analogies, among other
abstract reasoning tasks
\citep{webbEmergentAnalogicalReasoning2023, hanInductiveReasoningHumans2023, mirchandaniLargeLanguageModels2023, geigerRelationalReasoningGeneralization2023a}.

These empirical results highlight two significant points about the
relevance of deep learning to cognitive science. First, modern DNNs have
fulfilled the promise of older connectionist models in matching human
performance on many tasks probing core aspects of cognition. Second,
emerging evidence from interventional studies suggests that DNNs achieve
human-like performance on these tasks through mechanisms that differ in
nontrivial ways from those postulated by classical architectures
\citep{millierePhilosophicalIntroductionLanguage2024a}. Whether these
mechanisms are similar to those of human cognition remains an open
question that should be explored by experimentalists and philosophers of
cognitive science in tandem. At the very least, these findings suggest
that the classicist alternative to connectionism is no longer the ``only
game in town'' \citep{fodorLanguageThought1975} -- if it ever was.

The progress of DNNs has important implications for many other ongoing
issues in philosophy and cognitive science; I will briefly highlight two
that have attracted a lot of attention recently. The first pertains to
the ``grounding problem'' originally coined by
\citet{harnadSymbolGroundingProblem1990}: How can symbol-manipulating
systems have representations that are intrinsically connected to the
worldly referents of the symbols they manipulate? Without securing such
connection, it seems that computational models of cognition would have
difficulty escaping the ``merry-go-round'' of symbols and connecting to
the real world. While the grounding problem originally targeted
classical symbolic systems, it applies \emph{mutatis mutandis} to neural
networks used in natural langauge processing, such as LLMs, that
manipulate linguistic tokens. However,
\citet{molloVectorGroundingProblem2023} argue in light of philosophical
theories of representation that LLMs can in fact acquire world-involving
functions that secure norms of representational correctness relative to
the referrents of linguistic items.

Another hotly debated issue concerns the relevance of LLMs to
theoretical linguistics and theories of language acquisition. A wealth
of evidence from targeted experiments in computational linguistics
suggests that LLMs acquire sophiscated syntactic knowledge
\citep{linzenSyntacticStructureDeep2021a, pavlickSemanticStructureDeep2022b}.
This knowledge is reflected in the overall convergence of their
predictions with human grammaticality judgements regarding minimal pairs
of sentences that differ only with respect to some target linguistic
property \citep{warstadtBLiMPBenchmarkLinguistic2020}. LLMs'
representations of syntactic features can also be linearly decoded from
the activations of the network and manipulated with predictable effects
on behavior
\citep{ravfogelCounterfactualInterventionsReveal2021, belinkovProbingClassifiersPromises2022, haoVerbConjugationTransformers2023}.
These results call for an examination of the potential for deep learning
to inform linguistic theory
\citep{linzenWhatCanLinguistics2019a, dupreWhatCanDeep2021, baroniProperRoleLinguistically2022}.
In particular, it has been argued that LLMs challenge core tenets of
generative linguistics, on which statistical approaches to language
modelling relying on linear string order cannot account for the
hierarchical structure dependence of syntactic competence
\citep{chomskySyntacticStructures1957, everaertStructuresNotStrings2015, contreraskallensLargeLanguageModels2023, piantadosiModernLanguageModels2023}.
Most LLMs are exposed to a vastly greater quantity of words than
children during their learning phase \citep{frankBridgingDataGap2023},
which typically limits their relevance to debates about linguistic
nativism. Nonetheless, ongoing efforts to train LLMs in developmentally
plausible learning scenarios may vindicate their usefulness as model
learners that can constrain theories of language acquisition
\citep{warstadtWhatArtificialNeural2022, milliereLanguageModelsModelsforthcoming}.

\hypertarget{sec-methodological-issues}{%
\section{Methodological issues}\label{sec-methodological-issues}}

With the performance of DNNs improving across a range of linguistic and
cognitive tasks, the need for robust methods to evaluate DNNs and
compare them with humans under similar conditions is becoming more
pressing. Methodological insights from the philosophy of cognitive
science can inform evaluation practices in deep learning.

Behavioral evaluations based on benchmarks face challenges that
increasingly limit their usefulness in the age of LLMs. New benchmarks
tend to saturate very rapidly, although the best-performing models may
still exhibit failures modes in the target domain
\citep{kielaDynabenchRethinkingBenchmarking2021, ottMappingGlobalDynamics2022}.
The perverse incentive of ``SOTA-chasing'' -- pursuing
\emph{state-of-the-art} status on benchmark leaderboards -- can lead to
the exploitation of proxy metrics that diverge from the underlying
evaluation targets, in accordance with Goodhart's Law
\citep{manheimCategorizingVariantsGoodhart2018, gururanganAnnotationArtifactsNatural2018}.
Because LLMs are trained on internet-scale data, benchmark contamination
is also a common issue: test samples can easily leak into the training
data, leading to misleading improvements on standard evaluation metrics
\citep{zhouDonMakeYour2023}. Finally, the connection between latent
theoretical constructs and operational variables measured through
benchmarks is not always explicit or well-supported.

These challenges can be addressed through hypothesis-driven experiments
that incorporate best practices inspired by cognitive science. For
example, researchers can use novel stimuli to avoid data contamination;
use minimal stimuli (such as minimal pairs of sentences) to avoid
confounds; control the training data (by analogy with controlled rearing
experiments, e.g. \citet{leeControlledrearingStudiesNewborn2021}); use
multiple tasks to test the same capacity; collect multiple model
responses for each test item; and use appropriate control conditions
\citep{frankLargeLanguageModels2023}. Consider for example the question
whether LLMs are capable of acquiring a Theory of Mind (ToM). Using a
classic false-belief task, \citet{kosinskiTheoryMindMight2023} suggests
that GPT-3 exhibits ToM reasoning comparable to 9-year-old children.
However, performance success on a single task does not provide robust
evidence of the underlying competence. Indeed,
\citet{ullmanLargeLanguageModels2023} shows that minor conceptual task
variations, which maintain the core demands for false belief inference,
reveal the model's lack of abstract reasoning about mental states.
Patterns of performance can only constrain inferences about competence
given well-supported background assumptions about the measuring
instrument, measuring conditions, and target system.

The distinction between performance and competence cuts both ways
\citep{firestonePerformanceVsCompetence2020}. When comparing performance
across humans and DNNs, it is crucial to create adequately matched
testing conditions. For example,
\citet{lakretzCanTransformersProcess2022} suggest that Transformer
models like GPT-2 are fundamentally limited in their capacity to process
long-range recursive nesting compared to humans. In their experiment,
however, human subjects received substantial training with examples,
instructions, and feedback -- while GPT-2 was tested ``zero-shot''
without equivalent context. \citet{lampinenCanLanguageModels2023} shows
that after adding context analogous to human training, LLMs actually
perform \emph{better} than humans even on the most challenging
conditions. These examples highlights how careful one should be in
interpreting both success and failure modes of DNNs on tasks originally
designed for humans. Methodological principles from comparative and
developmental psychology can help mitigate comparative biases in
experimental design and analysis
\citep{bucknerBlackBoxesUnflattering2021}.

\hypertarget{conclusion}{%
\section{Conclusion}\label{conclusion}}

The progress of deep learning over the past decade has been more
significantly driven by engineering achievements than by theoretical
insights from cognitive science. This certainly does not mean that it is
irrelevant to cognitive science; nor does it mean that cognitive science
has nothing to contribute to deep learning research in return. Modern
DNNs do not merely mark incremental improvements over older neural
networks models, but represent a turning point for the connectionist
programme. While they still fall short of human cognitive competence in
various ways, and show noteworthy dissimilarities with human biases and
developmental trajectories, they also demonstrate an unprecedented
convergence with human performance on many long-standing challenges --
many of which were once widely thought to be hard limitations of
connectionist architectures. More than ever, neural networks show
promise as \emph{cognitive models} that can be systematically studied
and manipulated by scientists in carefully controlled conditions to
enable surrogative reasoning about core aspects of cognition. More than
ever, the need for rigorous evaluations of neural networks requires
interdisciplinary insights, particularly when it comes to
theoretically-informed comparisons with humans on linguistic and
cognitive tasks. Philosophers of cognitive science have much to
contribute to both theoretical and methodological issues raised by deep
learning.

\bibliography{bibliography}

\end{document}